\documentclass[letterpaper, 10 pt, journal, twoside]{IEEEtran}
\usepackage{graphicx}
\graphicspath{{images/}}
\usepackage{amsmath}
\usepackage[caption=false,font=footnotesize]{subfig}
\usepackage[thinlines]{easytable}
\usepackage{cite}
\usepackage{textcomp}
\usepackage{stfloats}
\usepackage{color}
\usepackage{booktabs}
\usepackage{textcomp}
\usepackage{microtype} % Improved Tracking and Kerning
\usepackage{hyperref}
\hypersetup{
    colorlinks=true,
    linkcolor=black,
    urlcolor=magenta,
    }
\usepackage{ccicons}  % Cite your images correctly!
\usepackage{xspace}

\usepackage{tabularx}
\usepackage{amssymb}
\usepackage{multirow}
\usepackage{verbatim}
\usepackage{booktabs}

\renewcommand{\baselinestretch}{0.97}

\makeatletter
\DeclareRobustCommand\onedot{\futurelet\@let@token\@onedot}
\def\@onedot{\ifx\@let@token.\else.\null\fi\xspace}

\makeatother

\usepackage{lipsum}

\usepackage[dvipsnames]{xcolor}     % Colours
\usepackage{amsfonts}               % Math fonts
\usepackage{mathrsfs}
\let\labelindent\relax             % this command is used in ieee and enumitem
\usepackage{enumitem}              % More control over type of bullet / number
\usepackage{algorithm}             % Algorithms. Need the next line too.
\usepackage[noend]{algpseudocode}  % Pseudocode setup
\usepackage{ifthen}
\usepackage[normalem]{ulem}        % For strikethrough
\usepackage[capitalise]{cleveref}
\def\arxiv{1}

\long\def\xspace{\mathcal{X}}

\usepackage{soul}

\long\def\revision#1{\textcolor{black}{#1}}
\long\def\revisionmove#1{}
\long\def\revisionmovehere#1{\textcolor{black}{#1}}
\long\def\revisionremove#1{}
\title{
Stein Variational Belief Propagation for \\ Multi-Robot Coordination
}

\author{Jana Pavlasek$^{1}$, Joshua Jing Zhi Mah$^{1}$, Ruihan Xu$^{1}$, Odest Chadwicke Jenkins$^{1}$, and Fabio Ramos$^{2}$
\ifthenelse{\equal{\arxiv}{1}}{% No RA-L info on arxiv.
}{%
\thanks{Manuscript received: October 12, 2023; Revised January 17, 2024; Accepted February 20, 2024. 
This paper was recommended for publication by Editor M. Ani Hsieh upon evaluation of the Associate Editor and Reviewers' comments.
This work was supported in part by Ford Motor Company, J.P.~Morgan AI Research, Amazon, the Alfred P. Sloan Foundation, and an NSERC doctoral fellowship.} %Use only for final RAL version
}
\thanks{$^{1}$J.~Pavlasek, J.~J.~Z.~Mah, R.~Xu, and O.~C.~Jenkins are with the Robotics Department, University of Michigan, Ann Arbor, USA
        {\tt\footnotesize \{pavlasek, joshmah, rhxu, ocj\}@umich.edu}}%
\thanks{$^{2} $F.~Ramos is with the NVIDIA Corporation, Seattle, USA \& School of Computer Science, University of Sydney, Sydney, Australia
        {\tt\footnotesize fabio.ramos@sydney.edu.au}}%
\ifthenelse{\equal{\arxiv}{1}}{% No RA-L info on arxiv.
\thanks{This work was supported in part by Ford Motor Company, J.P.~Morgan AI Research, Amazon, the Alfred P. Sloan Foundation, and an NSERC doctoral fellowship.}  % Put sponsorship down here on arxiv.
}{%
\thanks{Digital Object Identifier (DOI): see top of this page.} %Use only for final RAL version
}
}

\ifthenelse{\equal{\arxiv}{1}}{%
\markboth{}{}  % No header on arxiv.
}{%
\markboth{IEEE Robotics and Automation Letters. Preprint Version. Accepted March 2024}
{Pavlasek \MakeLowercase{\textit{et al.}}: Stein Variational Belief Propagation for Multi-Robot Coordination} 
}

\begin{document}

\maketitle
% UNCOMMENT FOR SUBMISSION
% \thispagestyle{empty}
% \pagestyle{empty}

% COMMENT FOR SUBMISSION
% \thispagestyle{plain}
% \pagestyle{plain}

\begin{abstract}
Decentralized coordination for multi-robot systems involves planning in challenging, high-dimensional spaces. The planning problem is particularly challenging in the presence of obstacles and different sources of uncertainty such as inaccurate dynamic models and sensor noise. 
In this \ifthenelse{\equal{\arxiv}{1}}{paper}{letter}, we introduce Stein Variational Belief Propagation (SVBP), a novel algorithm for performing inference over nonparametric marginal distributions of nodes in a graph.
We apply SVBP to multi-robot coordination by modelling a robot swarm as a graphical model and performing inference for each robot.
We demonstrate our algorithm on a simulated multi-robot perception task, and on a multi-robot planning task within a Model-Predictive Control (MPC) framework, on both simulated and real-world mobile robots.
Our experiments show that SVBP represents multi-modal distributions better than sampling-based or Gaussian baselines, resulting in improved performance on perception and planning tasks. Furthermore, we show that SVBP's ability to represent diverse trajectories for decentralized multi-robot planning makes it less prone to deadlock scenarios than leading baselines. 
\end{abstract}


\section{Introduction}
% Introduction

\IEEEPARstart{M}{ulti-robot} coordination is an essential capability for applications involving teams of robots, such as industrial robots, delivery vehicles, and autonomous cars. Planning for multi-robot systems is challenging due to the high-dimensionality introduced by a large number of agents.
Decentralized algorithms enable 
each robot to perform local computations using information from neighboring robots. This distributed approach is well-suited to multi-robot systems since it involves solving lower-dimensional, local problems compared to the expensive high-dimensional centralized approach. 

\begin{figure}[t]
    \centering
    \includegraphics[width=\linewidth]{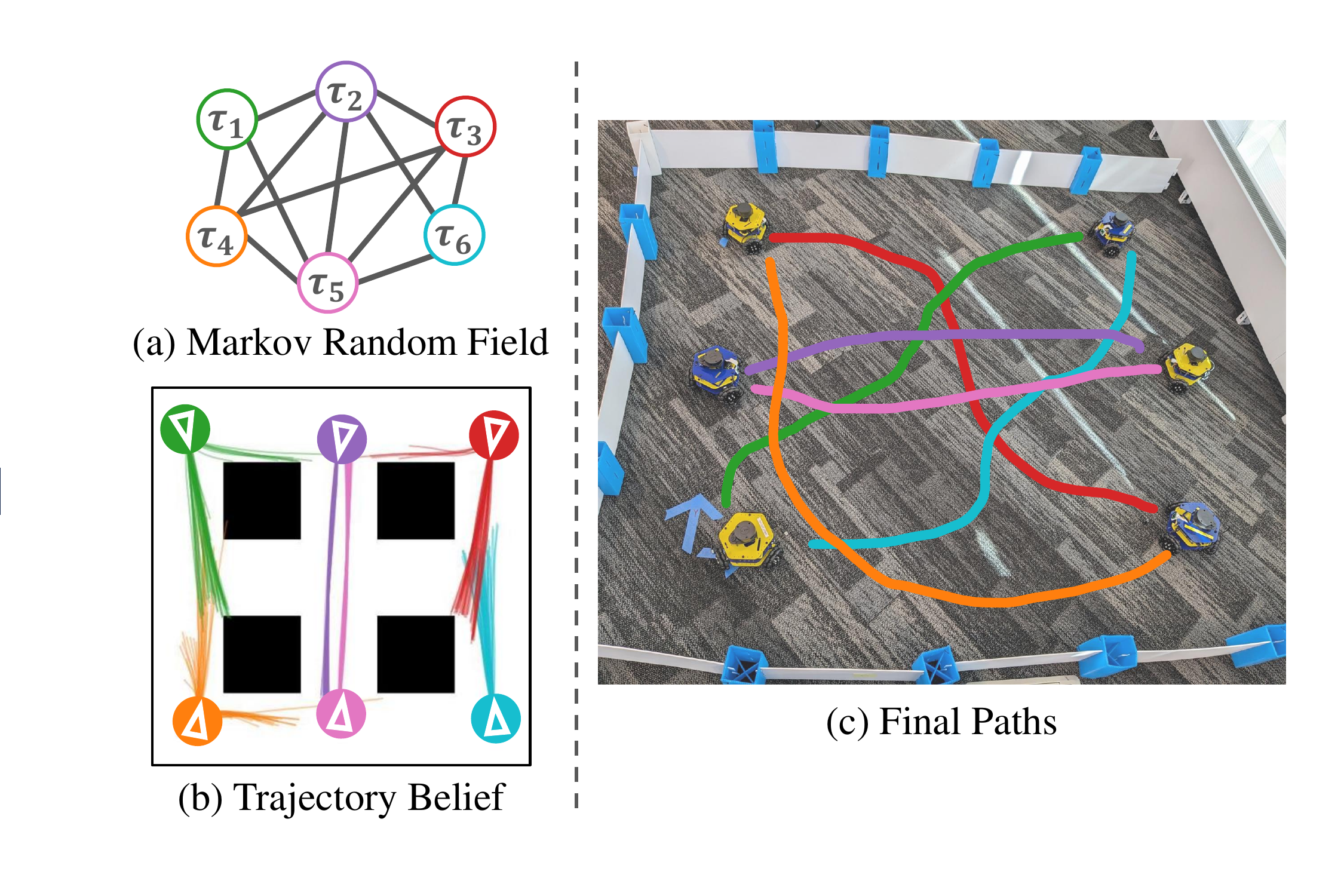}
    \caption{Stein Variational Belief Propagation (SVBP) computes marginal trajectory distributions for each robot in a multi-robot system. SVBP represents the relationships between robots as a Markov Random Field (a) and maintains multi-modal distributions over each robot trajectory (b). 
    Example final trajectories for each robot are shown in (c).
    }
    \label{fig:pitch}
\end{figure}

Decentralized control algorithms~\cite{fiorini1998motion, van2011reciprocal} are prone to deadlock scenarios which arise from the multi-modality of the solutions that each robot must consider. 
Considering multiple possible trajectories as a \textit{distribution} allows us to represent diverse solutions~\cite{pmlr-vR4-attias03a, toussaint2009robot}. This ability lends added robustness in dynamic environments, such as those with multiple mobile agents.
We therefore consider the problem of multi-robot coordination as a probabilistic inference problem.
We represent the robot swarm as a graphical model, where each robot is a node in the graph, and edges connect robots in communication range~\cite{schwertfeger2007multi, tolstaya2020learning, patwardhan2022distributing}. 
This representation enables distribution of possible trajectories for each robot to be inferred via graphical inference (see \cref{fig:pitch}).

We propose Stein Variational Belief Propagation (SVBP), an algorithm for performing probabilistic inference on a Markov Random Field (MRF) through message passing, and demonstrate its applicability to multi-robot coordination.
SVBP employs Stein Variational Gradient Descent (SVGD)~\cite{liu2016stein} to infer marginal posterior distributions as a set of particles through nonparametric belief propagation~\cite{sudderth2010nonparametric, isard2003pampas}.
Leveraging SVGD enables effective representation of multi-modal distributions, mitigating mode collapse compared to sampling-based methods. %, and
Our formulation extends SVGD to graphical models by leveraging the particle message update rules from Particle Belief Propagation (PBP)~\cite{ihler2009particle}. 
In contrast to SVGD, SVBP approximates the marginals rather than the full posterior, and can therefore scale to higher dimensional problems. 
The resulting algorithm is highly parallelizable since the particles are deterministically updated using gradient information, making it well-suited to efficient implementation on a GPU.

We demonstrate our approach on two applications: a simulated multi-robot perception task, and a multi-robot Model-Predictive Control (MPC) task, both in simulation and on a real-world mobile robot swarm.
We demonstrate how these problems can be formulated as MRFs~\cite{schwertfeger2007multi,butterfield2009modeling} and solved via SVBP. 
The belief propagation framework
enables multi-hop information to be passed through the graph while only passing messages between immediate neighbors.
The perception experiments show that SVBP can maintain multi-modal belief distributions in uncertain environments, leading to lower localization error compared to baselines.
The planning experiments demonstrate that SVBP is more resilient to deadlock scenarios, and produces smoother trajectories resulting in faster time-to-goal.
Our robot experiments show that our SVBP controller is robust to noisy localization and dynamics and asynchronous message passing. 
Video results are available %
\ifthenelse{\equal{\arxiv}{1}}{}{in the media attachment and} 
at: \href{https://progress.eecs.umich.edu/projects/stein-bp/}{https://progress.eecs.umich.edu/projects/stein-bp}.

\section{Related Work}
% Related Work

\textit{Decentralized} multi-robot coordination algorithms are those in which each robot executes a controller to satisfy individual objectives considering local information from neighbors. This technique is highly scalable to large and dynamic swarms. 
Optimal Reciprocal Collision Avoidance (ORCA)~\cite{van2011reciprocal}, a variant of velocity obstacles~\cite{fiorini1998motion}, demonstrates real-time collision avoidance for thousands of agents with independent objectives but are highly prone to deadlock scenarios. We focus on decentralized Model-Predictive Control and graphical approaches in this section and refer the reader to existing surveys~\cite{gerkey2004formal,rossi2021multi} for broader coverage. 

\textbf{Multi-robot coordination with graphical models:}
Probabilistic graphical models present a natural formulation for decentralized multi-robot coordination, whereby individual robots are represented by nodes in a graph and edges connect communicating robots~\cite{schwertfeger2007multi}.
This formulation has been used to solve for robot localization and control with Gaussian Belief Propagation~\cite{patwardhan2022distributing}. 
Graphical representations have also been used to learn factors for robot control via graph neural networks~\cite{tolstaya2020learning, tolstaya2021learning}. This technique requires expert trajectory demonstrations from a centralized controller for training.

\textbf{Multi-robot Model Predictive Control:}
Decentralized model-predictive control (DMPC) has been applied to multi-agent collision avoidance problems~\cite{morgan2014model, ong2015coop, van2017dmpc, dai2017distributed, luis2020online}. 
By planning over a horizon, these techniques mitigate deadlock scenarios issues but introduce complexities due to the higher dimensionality introduced. These works consider the problem of finding a single trajectory solution.
In work most similar to ours, Patwardhan et al. use Gaussian Belief Propagation for collision avoidance in multi-robot planning~\cite{patwardhan2022distributing}. This method restricts the trajectory distributions to Gaussian forms, and requires all factors to be \revision{linearized about an estimate}. % linear Gaussian
In contrast, our approach can be used with any differentiable factor and uses a more flexible nonparametric distribution.

\section{Background}

\subsection{\revision{Belief Propagation}}

% \marginnote{\#M.2}[-1.5cm]
Let $G=(V, E)$ denote a Markov Random Field (MRF) with nodes $V$ and edges $E$. Let $\mathcal{X} = \{x_s \mid s\in V\}$ denote the set of all hidden nodes in the graph, and $\mathcal{Z}= \{z_s \mid s\in V\}$ denote the observed nodes corresponding to each hidden node.
The joint probability of the graph $G$ can be expressed as a product of its clique potentials:
\begin{equation}
\label{eq:jointprob}
    p(\mathcal{X}, \mathcal{Z}) \propto \prod_{(s,t)\in E}\psi_{st}(x_s, x_t) \prod_{s \in V}\phi_{s}(x_s, z_s).
\end{equation}
The function $\psi_{st}$ is the \textit{pairwise potential}, describing the correspondence between neighboring nodes, and $\phi_{s}$ is the \textit{unary potential}, describing the correspondence of a hidden variable $x_s$ with the observed variable $z_s$. 
Note that in the following equations, we omit the observation, $z_s$, from the unary potential $\phi_s$ for brevity. 

Belief propagation estimates the marginal posterior distribution of a hidden node $s$ using the following equation:
% Sum Product BP solves for the marginal beliefs at each hidden node in an MRF, which is described by Equation~(\ref{eq:jointprob}). The belief for node $s$ is given by:
\begin{equation}\label{eq:bp_sum_bel}
    p(x_s\mid\mathcal{Z}) \propto \phi_s(x_s) \prod_{t\in \rho(s)} m_{t\to s}(x_s)
\end{equation}
where $\rho(s)$ denotes the neighbors of $s$. The message from node $t$ to node $s$, $m_{t\to s}$, is defined as:
\begin{equation}\label{eq:bp_sum_msg}
    m_{t\to s}(x_s) = \int_{x_t} \phi_{t}(x_t) \psi_{ts}(x_t, x_s) \prod_{u\in \rho(t)\setminus s} m_{u\to t}(x_t) \; dx_t
\end{equation}
\revisionremove{In many cases, the integral in the above equation is intractable. Sampling-based approximations are a common way to circumvent this issue.}
% Note that we omit the observation, $z_s$, from the unary potential $\phi_s$ for brevity. 
% \pe{maybe put this before (2)}. \rx{Agree, because we mention BP in (2). The knowledge will build up more smoothly from a perspective of non-expert in this field}
% 
\revision{%
Belief propagation provides exact marginals for tree structured graphs. 
For graphs with loops, messages can be computed iteratively.
% , in which an updated message is computed using the estimate for $m_{u\to t}(x_t)$ from the last iteration. 
This approach is proven effective in practice~\cite{murphy1999loopy}.  
We refer the reader to Wainwright and Jordan~\cite{wainwright2008graphical} for further details on message passing techniques.
% Each message passing algorithm is described in Algorithm~\ref{alg:msg_pass}.
} %\jp{TODO: Move the reference to the loopy version of BP to here.}

\revisionmovehere{A number of belief propagation algorithms have been proposed in the literature.
Gaussian Belief Propagation (GaBP) is an efficient algorithm when the node distributions and their corresponding factors can be represented as Gaussian~\cite{weiss1999correctness, davison2019futuremapping}. This method enables efficient computation and has been shown to be effective for multi-robot collision avoidance and localization~\cite{patwardhan2022distributing, murai2022robot}. However, many applications in robotics are complex and multi-modal, and cannot be fully represented by unimodal Gaussian uncertainty.}

% \subsection{Particle Belief Propagation}

% \begin{algorithm}[t]
% \caption{\revision{Message update algorithm}}\label{alg:msg_pass}
% \begin{algorithmic}
% \Procedure{update\_messages}{$x_s, G$}
% \For{$(s, t) \in E$}
% \State Sample $x_t^{(j)} \sim W_t$, $j=1\dots M$
% \For{$u \in \rho(t)\setminus s$}
% \State $\hat{m}_{u\to t}(x_t^{(j)}) \gets \Call{update\_messages}{x_t}
% $
% \EndFor
% \State Update $\hat{m}_{t\to s}(x_s^{(i)})$ \Comment{Eq.~\eqref{eq:pbp_msg}}
% \EndFor
% \EndProcedure
% \Procedure{update\_messages\_loopy}{G}
% \State Initialize messages $(m_{t\to s}, m_{s\to t})$ for $(s, t) \in E$
% \For{$k = 1, \dots, K$}
% \For{$(s, t) \in E$}
% \State Sample $x_t^{(j)} \sim W_t$, $j=1\dots M$
% \For{$u \in \rho(t)\setminus s$}
% \State $\hat{m}_{u\to t}(x_t^{(j)}) \gets m^{k - 1}_{u\to t}(x_t^{(j)})
% $
% \EndFor
% % \State Sample $x_s^{(j)} \sim W_s$, $j=1\dots M$
% \State Update $\hat{m}^k_{t\to s}(x_s^{(i)})$ \Comment{Eq.~\eqref{eq:pbp_msg}}
% \EndFor
% \EndFor
% \EndProcedure
% \end{algorithmic}
% \end{algorithm}

\subsection{Particle Belief Propagation}

\revisionmovehere{Nonparametric Belief Propagation (NBP)~\cite{sudderth2010nonparametric, isard2003pampas} represents distributions nonparametrically as mixtures of Gaussians and are well-suited to cases where the integral in Eq.~\eqref{eq:bp_sum_msg} is intractable. NBP algorithms involves expensive product operations between mixture distributions. NBP has been applied to robotic perception of articulated objects, using an efficient sampling-based message product technique~\cite{Desingh2019pmpnbp}, learned unary factors~\cite{pavlasek2020parts}, and end-to-end learned factors~\cite{Opipari:2023:DNBP:3592762}. While these methods enable complex representations of belief distributions, they rely on expensive sequential sampling operations.}
% Particle Belief Propagation (PBP) was proposed by Ihler and McAllester~\cite{ihler2009particle}. Consider the sum-product formulation of belief propagation, where the messages and belief are defined in Equations (\ref{eq:bp_sum_bel}) and (\ref{eq:bp_sum_msg}) respectively.

Particle Belief Propagation (PBP) defines a sampling-based algorithm for computing the messages in Eq.~\eqref{eq:bp_sum_msg} for cases where the integral is intractable due to the complexity of the state space~\cite{ihler2009particle}. PBP represents the belief at each node with a set of $N$ particles, $\{x_s^{(i)}\}_{i=1:N}$. Given samples \revision{from node $t \in \rho(s)$,} $\{x_t^{(i)}\}_{i=1:M}$ drawn from a candidate distribution $W_t$, PBP defines the approximate message:
\begin{multline}\label{eq:pbp_msg}
    \hat{m}_{t\to s}( x_s^{(i)}) = \\
    \frac{1}{M} \sum_{j=1}^M \frac{\phi_{t}(x_t^{(j)})}{W_t( x_t^{(j)} )} \psi_{ts}(x_t^{(j)}, x_s^{(i)}) \prod_{u\in \rho(t)\setminus s} m_{u\to t}(x_t^{(j)}).
\end{multline}
This message definition is used to draw samples from the marginal posterior, $p(x_s\mid \mathcal{Z})$, using importance sampling. 

\revisionmovehere{PBP relies on the definition of a sampling distribution, $W_t$, which later work proposed to estimate via expectation maximization~\cite{lienart2015epbp}. Importance sampling is prone to mode collapse, which has been mitigated by drawing from multiple sampling distributions~\cite{pacheco2014dpmp}. However, current solutions to PBP require careful selection of sampling distributions and sequential sampling operations.
}
%In practice, a number of candidate distributions can be chosen, including the belief estimate itself.

\subsection{\revision{Stein Variational Inference}}

% \marginnote{\#2.2}%[-5cm]
\revision{
Stein Variational Inference is an algorithm for approximating a distribution $p(x)$ using a candidate distribution in the form of a set of particles, $q(x) = \{x^{(i)}\}_{i=1:N}$. Stein variational gradient descent (SVGD)~\cite{liu2016stein} employs gradient-based optimization over the particle set to minimize the kernelized Stein discrepancy~\cite{liu2016ksd} between the true density and a candidate density represented by the particle set.
SVGD is an iterative algorithm which applies the following update to each particle $i$ at iteration $k$:
\begin{align}
    x^{(i)}[k] &\leftarrow x^{(i)}[k-1] + \epsilon \gamma(x^{(i)}[k-1]) \\
    \gamma(x) &= \frac{1}{N} \sum_{j=1}^N \kappa(x^{(j)}, x) \nabla_{x^{(j)}} \log p(x^{(j)}) + \nabla_{x^{(j)}} \kappa(x^{(j)}, x)\label{eq:stein-op}
\end{align} 
where $\kappa(x^{(j)}, x)$ is a kernel function between particles. 
% The above equations are applied for a fixed number of iterations or until convergence. 
We can interpret the first term inside the summation of Eq.~\eqref{eq:stein-op} as an attractive force that moves particles according to the gradient of the log-density, while the second term acts as a repulsive term keeping particles from collapsing to a single point estimate. Thus SVGD leverages parallel gradient-based optimization to generate a diverse set of samples more efficiently than Markov chain Monte Carlo (MCMC) samplers. %~\cite{williams2018information}. 
}

\revisionmovehere{%
SVGD has proven useful in a number of robotic applications in recent years, including control, planning, and point cloud matching~\cite{lambert2020steinmpc, barcelos2021dual,9661415,9811656}. SVGD has been applied to graphical models to approximate joint distributions using kernels over local node neighborhoods~\cite{wang2018stein} and conditional distributions over nodes~\cite{zhuo2018message}. Both these methods rely on the conditional independence structure of MRFs and as such only pass messages between immediate neighbors in the graph. In contrast, our proposed method computes the \textit{marginal} beliefs over nodes using belief propagation, which involves passing messages through the whole graph.
}

\section{Stein Variational Belief Propagation}
\begin{algorithm}[t]
\caption{\revision{The SVBP Algorithm}}\label{alg:svbp}
\begin{algorithmic}
\Procedure{SVBP}{G, $\mathcal{Z}$} 
\State Initialize particles $\{x_s^{(i)}\}_{i=1:N}$ for $s \in V$
\For{$k = 1, \dots, K$}
\State Update messages $(m_{t\to s}, m_{s\to t})$ for $(s, t) \in E$
\For{$s \in V$}
\For{$i = 1, \dots, N$}
\State Compute $\gamma(x_s^{(i)})$ \Comment{Eq.~\eqref{eq:stein-op}}
\State $x_s^{(i)} \gets x_s^{(i)} + \epsilon \gamma(x_s^{(i)})$
\EndFor
\EndFor
\EndFor
\EndProcedure
\end{algorithmic}
\end{algorithm}

Given a Markov Random Field (MRF) $G=(V, E)$,
we seek to infer the marginal distribution of a node $s \in V$, $p(x_s)$, defined in Eq.~\eqref{eq:bp_sum_bel}.
We propose Stein Variational Belief Propagation (SVBP), an algorithm for inferring marginal beliefs in an MRF using SVGD. The marginal distribution is represented nonparametrically using a particle set for each node in the graph, $\{x_s^{(i)}\}_{i=1:N}$.
We use SVGD gradient updates to infer the density $p(x_s)$ for each node.
We define the posterior likelihood 
term in Equation~(\ref{eq:stein-op}) using the marginal belief from Eq.~\eqref{eq:bp_sum_bel}, $p(x_s)$, to obtain the SVBP likelihood gradient:
% , the update rule for the particles at node $s$ is:
\begin{multline}
    \nabla_{x_s} \log p(x_s) = \nabla_{x_s} \log \phi_s(x_s) \\ + \sum_{t\in \rho(s)} \nabla_{x_s} \log \revision{\hat{m}}_{t\to s}(x_s),\label{eq:sbp_grad}
\end{multline}
where $\revision{\hat{m}}_{t\to s}(x_s)$ is defined via the PBP message rule from Eq.~\eqref{eq:pbp_msg}. A distinct set of Stein particles represents the posterior belief at each node. 

The inference process using SVBP is described in Algorithm~\ref{alg:svbp}. Particles are first initialized based on the problem domain. At each iteration, messages are updated with Eq.~\eqref{eq:pbp_msg}. For each node, particles are updated using Eq.~\eqref{eq:stein-op}, computed by evaluating the gradients in Eq.~\eqref{eq:sbp_grad} and the kernel function. The process is repeated for $K$ iterations or until convergence.

SVBP provides several key advantages over other NBP techniques. First, it uses gradient-based, deterministic particle updates which can be efficiently parallelized on a GPU, without relying on sequential sampling operations. Second, SVGD is well-suited to multi-modal applications due to its ability to maintain diverse modes with fewer particles.
SVBP also defines the kernel function in Eq.~\eqref{eq:sbp_grad} over individual nodes in the graph. This makes SVBP well-suited to high-dimensional problems which can be represented as a graph.

\revision{\textbf{Computing Gradients:}}
% \marginnote{\#1.1}%[-5cm]
\revision{SVBP requires that potentials $\phi$ and $\psi$ are differentiable. The message gradients can be computed as follows:}
\revision{
\begin{align}
\nabla_{x_s} \log \hat{m}_{t\to s}(x_s) &= \frac{\nabla_{x_s} \hat{m}_{t\to s}(x_s)}{\hat{m}_{t\to s}(x_s)}\label{eq:sbp_msg_grad}\\
\begin{split}
\nabla_{x_s} \hat{m}_{t\to s}(x_s) &= \\
\frac{1}{M} \sum_{j=1}^M 
\frac{\phi_{t}(x_t^{(j)})}{W_t( x_t^{(j)})} & \left[ \nabla_{x_s}  \psi_{ts}(x_t^{(j)}, x_s) \right] \prod_{u\in \rho(t)\setminus s} \hat{m}_{u\to t}(x_t^{(j)})
\end{split}
\end{align}
}
\revisionmovehere{We note that the gradient update from Equation~(\ref{eq:sbp_grad}) only involves evaluating gradient information from immediate neighbors, since the messages $m_{u\to t}$ in Eq.~\eqref{eq:pbp_msg} are not a function of $x_s$. This enables efficient gradient updates since the algorithm only requires backpropagating through single-hop neighbors.}

\revision{\textbf{Sampling Distribution:}} In practice, we use the current belief of the neighboring node, $p(x_t)$, as the sampling distribution, $W_t$, where $p(x_t)$ is represented by Stein particles for node $t$ with equal weights. This enables efficient computation of the messages since it eliminates the need to run expensive sampling algorithms like MCMC, as originally proposed.% in the PBP algorithm. 

\revision{\textbf{Message Passing Schedule:}}
We employ a synchronous message passing scheme in which all messages are computed prior to updating each node belief. This enables efficient batch computations of factors and messages suitable for execution on a GPU. However, our algorithm can be employed with other message passing schedules.

\revision{\textbf{Selecting an Estimate:}}
In practice, multiple estimates exist for drawing an estimate from the particle set. In practice, we select the highest weighted estimate for the experiments described. The weights for the particles can be computed after convergence using Eq.~\eqref{eq:bp_sum_bel}, $w_s^{(i)} = \phi(x_s^{(i)})\prod_{t\in \rho(s)}  m_{t\to s}(x_s^{(i)})$, where the messages are computed using Eq.~\eqref{eq:pbp_msg}.

\section{SVBP for Multi-Robot Perception}

\begin{figure}[t]
    \centering
    \includegraphics[width=0.76\linewidth]{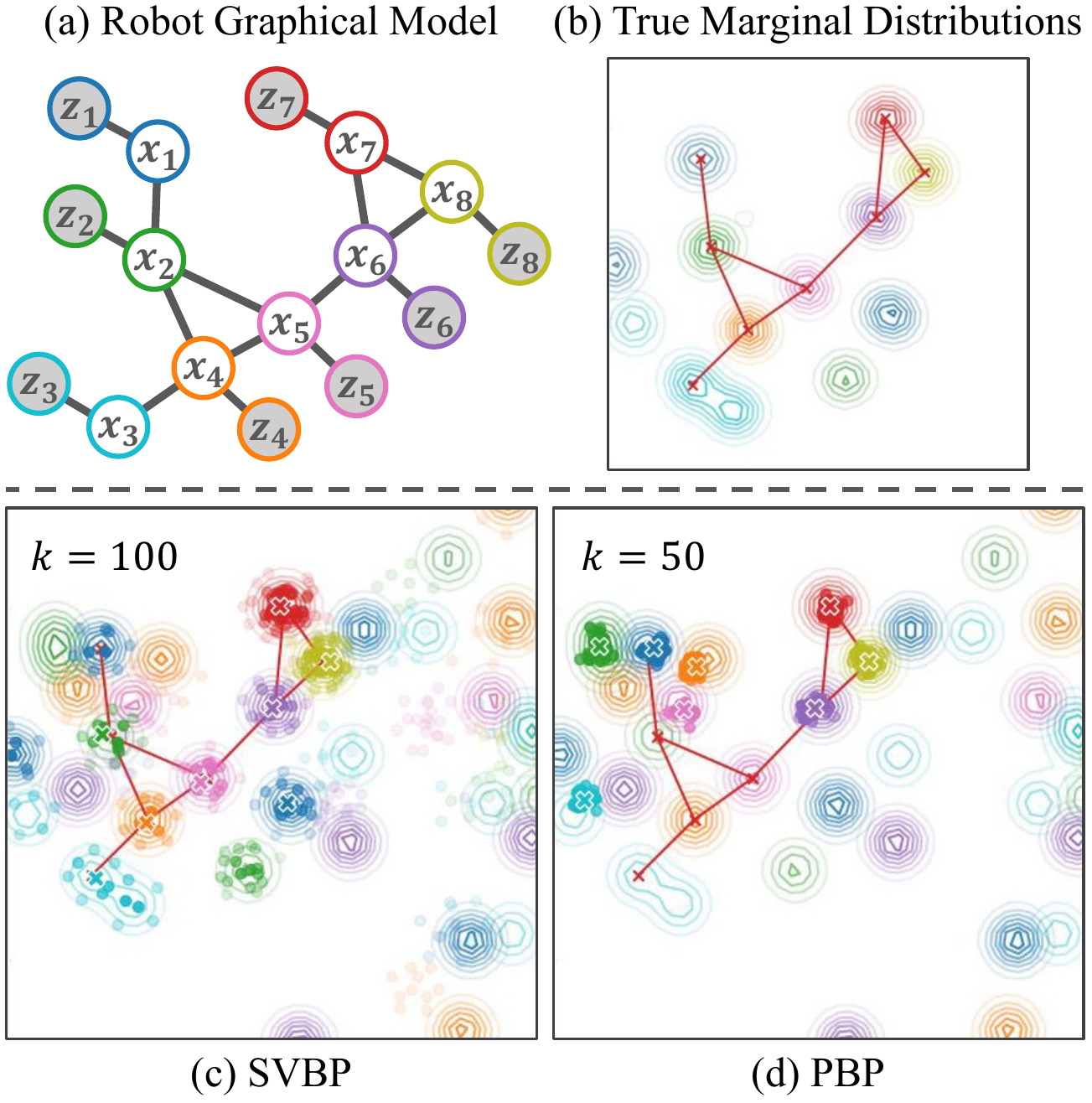}
    \caption{
    SVBP better represents the underlying distribution, avoiding mode collapse.  
    (a) Graphical model of the multi-robot perception problem. The position of each node is denoted $x_i$, and the corresponding observation is denoted $z_i$. (b) The approximate true marginals for the graph in (a) and the observation shown in (c, d).
    Qualitative results for SVBP (c) and PBP (d) at the final iteration ($k$). The red lines represent the true position of the nodes, and the colored `x' markers represent the maximum likelihood estimate for each node. Lower-weighted particles are shown with lower transparency. The distributions represent the noisy observations for each node of the corresponding color. Best viewed in color.}
    \label{fig:spider_qualitative}
\end{figure}

The first application on which we validate our algorithm is a simulated multi-robot perception experiment. The objective is to infer the belief, $p(x_s)$, over the robot's 2D position, denoted $x_s$, for each robot $s$ \revision{for a single timestep}. 
We consider the challenging case in which the observation 
for each agent is multi-modal. Specifically, the observation consists of a mixture of Gaussians which contains a component centered around the true position of the robot and randomly sampled noisy components. An example observation and the associated graphical model are shown in \cref{fig:spider_qualitative}. 
In addition to the observations, robots observe the displacement to neighboring robots \revision{within communication range}, creating edges in the graph (shown in red). The resulting marginal distributions for each robot are multi-modal, as shown in \cref{fig:spider_qualitative}(b). 
\revisionremove{We restrict the graph to a tree structure without loops for this problem. This experiment is a version of the articulated ``spider'' localization problem from the NBP literature}~\cite{isard2003pampas, Desingh2019pmpnbp, Opipari:2023:DNBP:3592762}.

The MRF in \cref{fig:spider_qualitative} requires the definition of the potentials in Eq.~\eqref{eq:jointprob}.
We define the unary potential for each robot to be the mixture of Gaussians corresponding to the robot observation. The pairwise potential is defined as a function of the observed translation $L_{st}$ between neighboring robots: 
\begin{equation}
    \psi_{ts}(x_t, x_s) = \exp \Big(-\alpha \big( \|x_s - x_t \| - L_{st} \big )^2 \Big ).
\end{equation}
where $x_s$ and $x_t$ are the 2D positions of neighboring robots \revision{and $\alpha$ is a user-selected coefficient}. 

\textbf{Baseline:} 
We implement Particle Belief Propagation as a baseline approach. We employ iterative importance sampling over the particles at each node, where each particle is weighted according to Eq.~\eqref{eq:bp_sum_bel} with the message definition of Eq.~\eqref{eq:pbp_msg}. We use the current particle set at each neighboring node as the candidate distribution for message computation, with equal weights, as in SVBP. 
We apply random noise at the beginning of each iteration. The same factor definitions and parameters are used for PBP and SVBP.

\begin{figure}[t]
    \centering
    \includegraphics[width=0.9\linewidth]{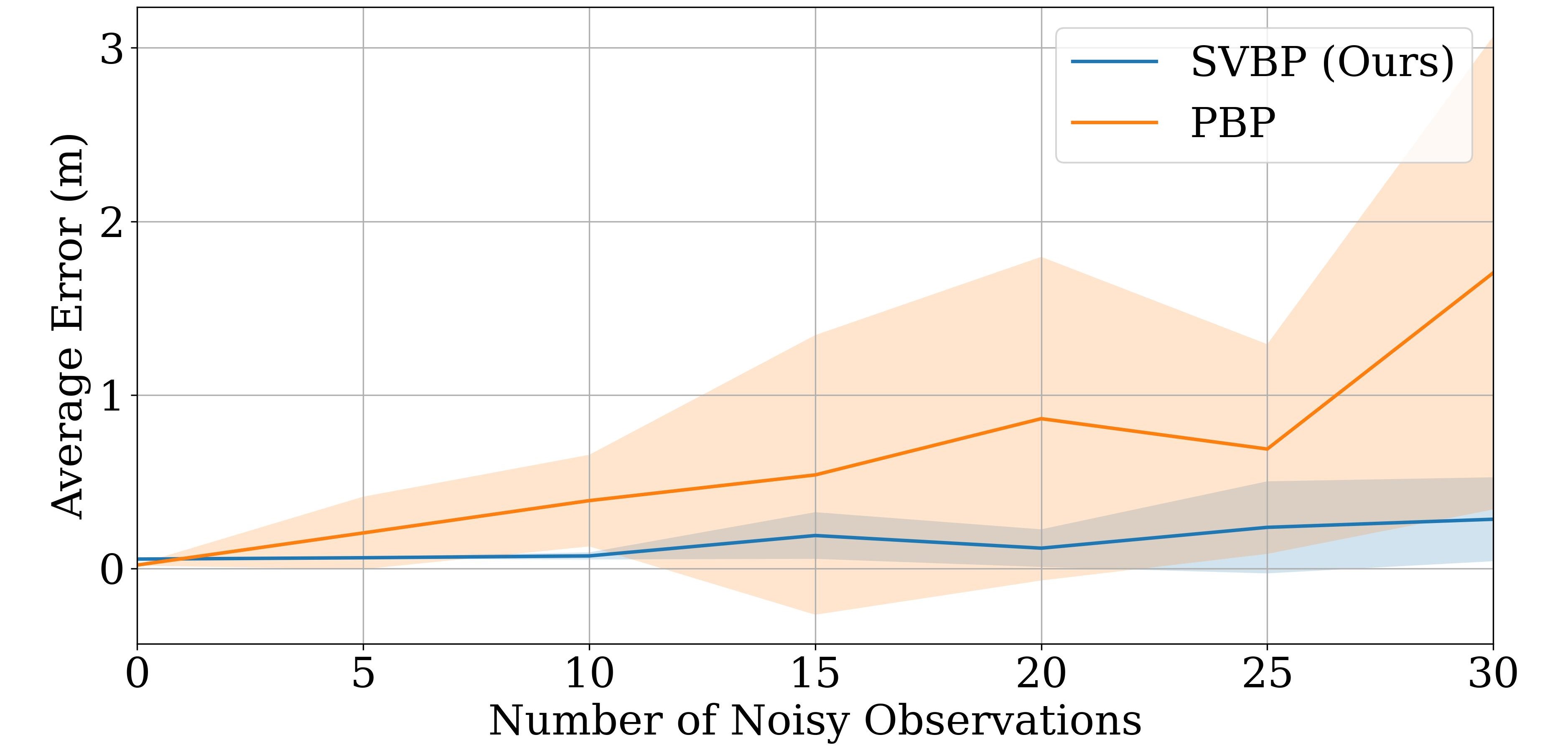}
    \caption{Average error for each node estimate for multi-robot localization. Results are shown for varying levels of noise, corresponding to the number of noisy components added to the observation.}
    \label{fig:spider_error}
\end{figure}

\subsection{Results}
For each run, \revision{the position of each robot is randomly selected such that the graph is connected for a radius of 2 meters. To form the observation, a component is added at the true location of the robot.} \revision{Noisy components with uniformly sampled means are then randomly assigned across nodes and added to the observation,}
making each observation a Gaussian mixture. \revision{Particles are initialized uniformly.} SVBP ran for 100 optimization 
iterations, and PBP ran for 50 iterations.
\revision{To generate an estimate for each node's position, we select the highest weighted particle.}

\revisionmovehere{The average error for 8 nodes over 10 runs for our SVBP algorithm against PBP is shown in \cref{fig:spider_error}.
The $x$-axis represents the total number of noisy Gaussian components added to the node observations.}
A visualization of the final belief distributions of SVBP and PBP for the highest noise observation is shown in \cref{fig:spider_qualitative}.
SVBP performs comparatively to PBP for low observation noise, but significantly outperforms PBP in noisy cases. We observed that PBP tends to converge quickly but was subjected to mode collapse which results in locally optimal estimates. In contrast, SVBP maintains multiple modes, making it more likely that the global solution is represented in the candidate particle set. 

\subsubsection{Comparison to True Marginals}
We hypothesize that SVBP better represents the true marginal distributions.
We perform an analysis of the particle distribution of each method compared to the true marginal beliefs. To obtain the true marginal beliefs, we run a Gibbs simulation~\cite{casella1992explaining} 
to sample from the marginal using the density from Eq.~\eqref{eq:bp_sum_bel}. To evaluate the integral for the message in Eq.~\eqref{eq:bp_sum_msg}, we employ Monte-Carlo integration over the full region of the observation with a high number of samples (1000). The ground truth sampled marginals are imperfect due to the sampling procedure but provide a reasonable baseline approximation. The visualization of the true marginal is shown in \cref{fig:spider_qualitative}(b).

\begin{figure}[t]
    \centering
    \includegraphics[width=0.88\linewidth]{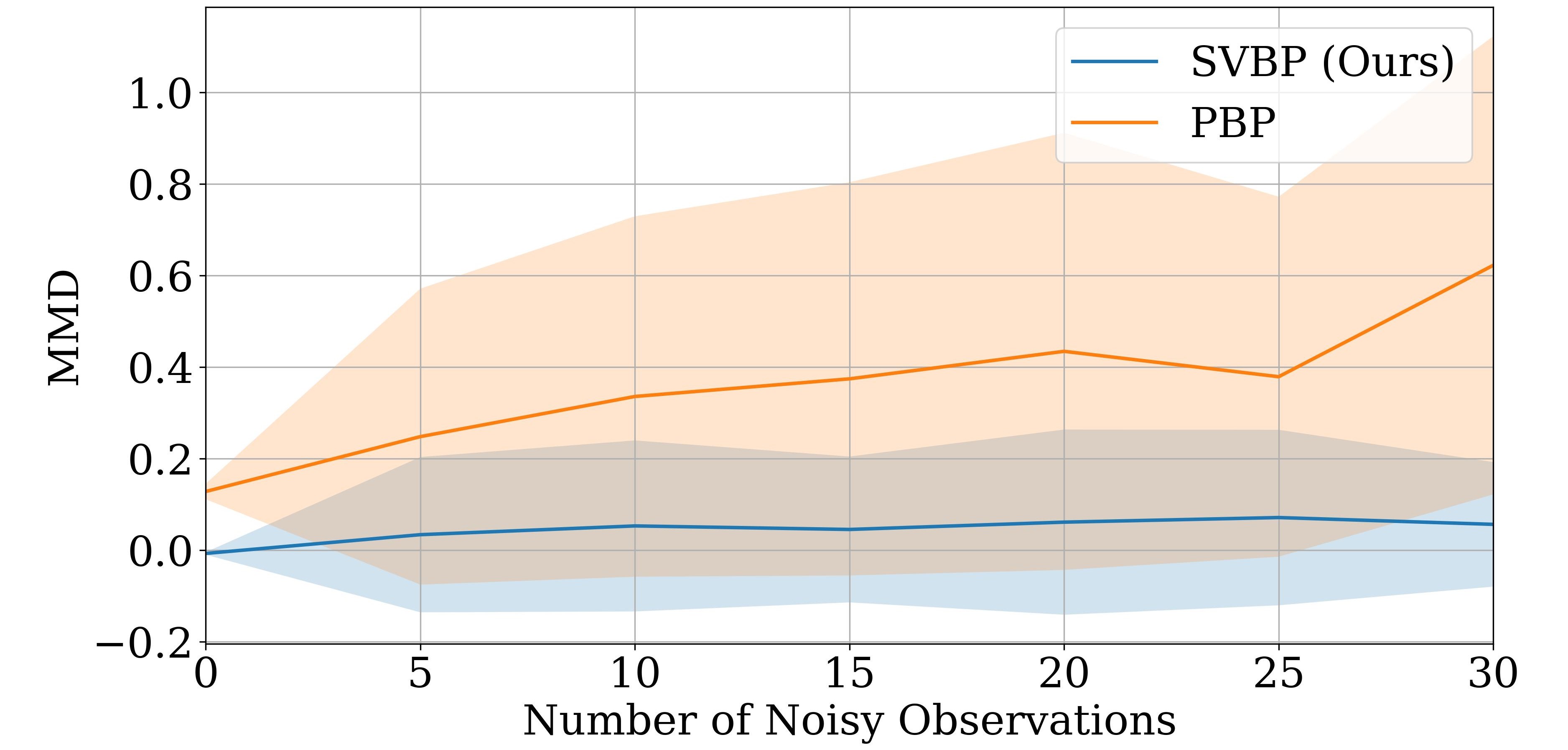}
    \caption{The average Maximum Mean Discrepency (MMD) between the samples from the true marginal distribution and the particle sets from SVBP and PBP. Both methods use 50 particles. }
    \label{fig:spider_mmd}
\end{figure}

We compute the kernelized Maximum Mean Discrepancy (MMD)~\cite{gretton2012kernel} between the sampled particle set and the belief particles for SVBP and PBP. The kernel bandwidth is chosen using the median heuristic over the ground truth sample set~\cite{garreau2017large}.
Results are shown in \cref{fig:spider_mmd}.
SVBP obtains a lower MMD than PBP consistently across noisy environments. 
We observe that some particles in SVBP get caught in local minima in very noisy cases in areas where the unary potential is high, as in \cref{fig:spider_qualitative}(c). These particles are easily detected as they have very low overall weights and could be reset in practice. We therefore do not include any particles with weights less than 1\% of the highest weight in the MMD computation. 
% We postulate that results could be further improved through more aggressive resetting schedules.

\begin{figure}[t]
    \centering
    \includegraphics[width=0.88\linewidth]{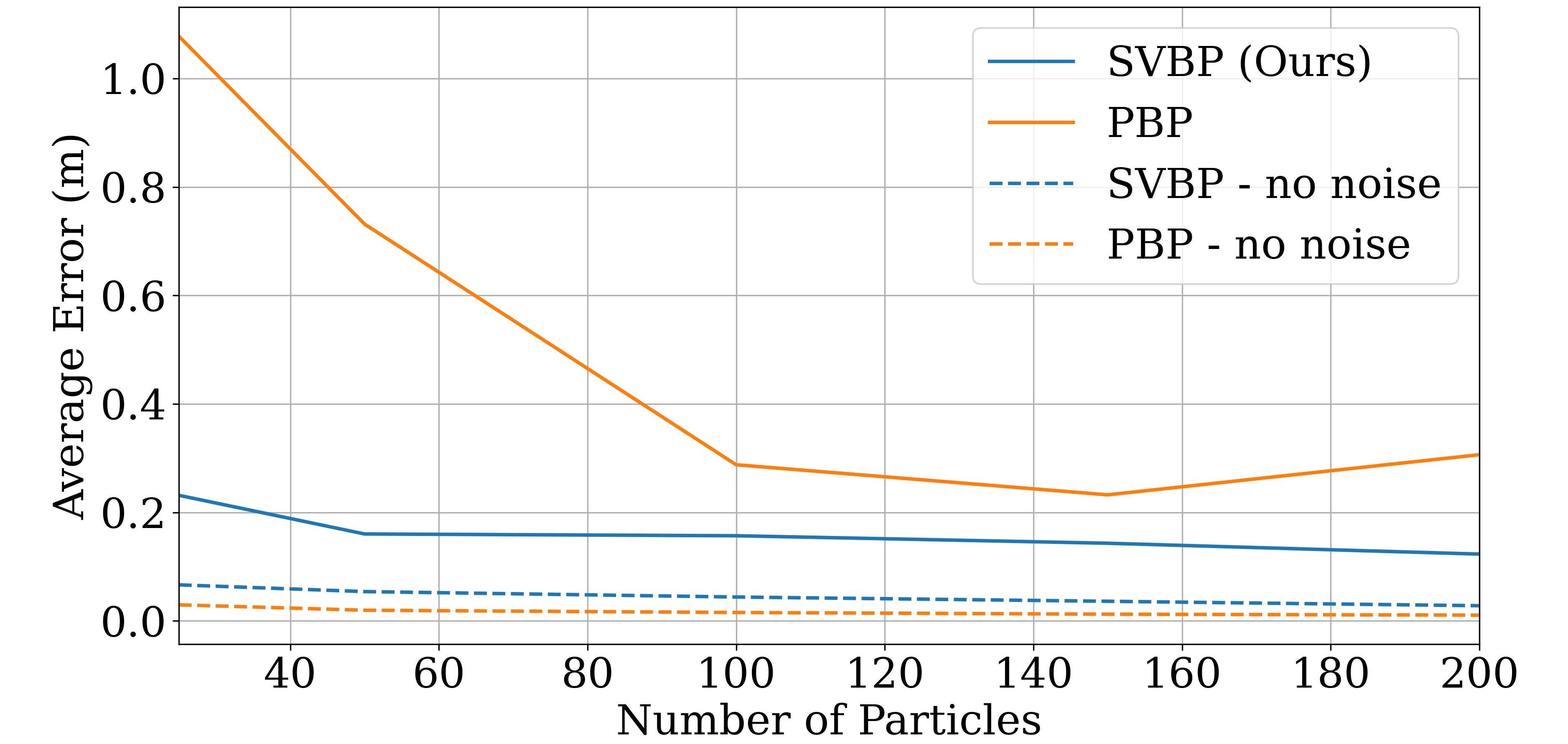}
    \caption{Average error for each node estimate for different numbers of particles. The solid lines correspond to experiments runs with noise added to the observation. The dashed lines correspond to experiments with no noise added to the observation.}
    \label{fig:spider_particles}
\end{figure}

\subsubsection{Analysis of Number of Particles}
We claim that SVBP can represent the marginal beliefs with fewer particles due to SVGD's ability to maintain modes of the distribution. We execute both SVBP and PBP with different particle set sizes and measure the average error across each node for the final estimate. The results are shown in \cref{fig:spider_particles}. 
For noisy environments, PBP benefits significantly when the size of the particle set is increased from 25 to 100, whereas SVBP finds a good estimate with only 25 particles. 
% \revisionremove{
% For environments with no noise, PBP has a slight advantage over SVBP since the particles are allowed to collapse to the true estimate, whereas SVBP particles might not converge to the true value due to the repulsive force. 
% This could be mitigated by performing gradient descent without the Stein update after convergence to improve the final estimate.
% }

% \section{Experiments \& Results}
\section{SVBP for Multi-Robot Planning}
% RESULTS

\begin{figure*}[t]
    \centering
    \includegraphics[width=0.8\linewidth]{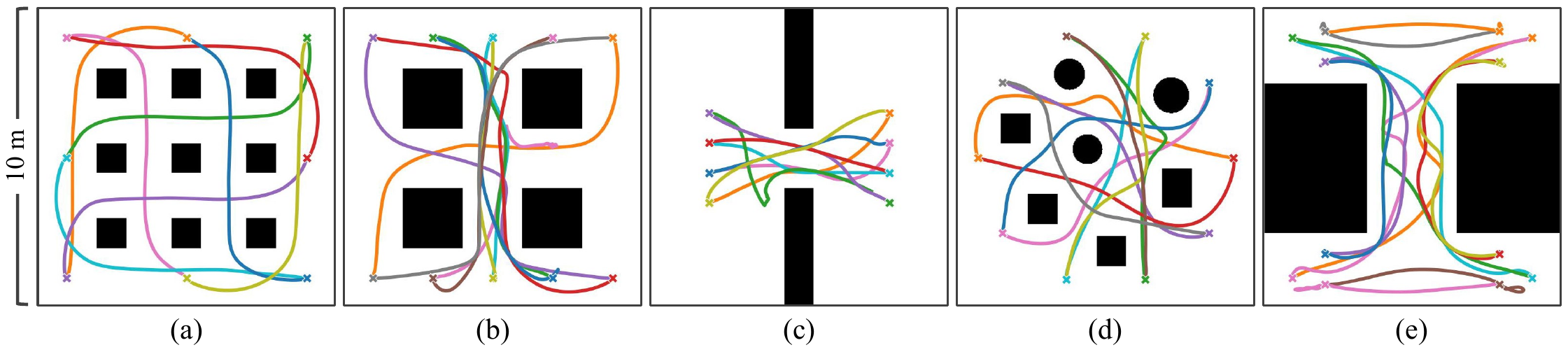}
    \caption{Testing environments for the multi-robot control experiments with randomly selected trajectories from SVBP. 
    The goal positions for each robot are marked with an `x'. Each environment is 10 meters by 10 meters.
    }
    \label{fig:swarm_envs}
\end{figure*}

Our second application involves decentralized Model Predictive Control (MPC) of a multi-robot system. Each robot must avoid obstacles and the other robots in its trajectory to the goal.  We run experiments both in a 2D planar navigation simulation and on a decentralized real robot system with realistic sensor and action noise. 

\subsection{Problem Formulation}

We consider the problem of finding a collision-free trajectory for each robot $s$, $\tau_s=\{u_{s,k}\mid 1 \leq k \leq T\}$, where $u_{s,k}$ are control commands for time $k$ over a fixed horizon $T$.
We take a planning as inference approach~\cite{pmlr-vR4-attias03a, toussaint2009robot} 
in which the nodes in the graph represent the trajectory distribution, $p(\tau_s)$ for each robot, and the edges in the graph represent robots in communication, as in \cref{fig:pitch}.
% We consider the trajectories to be a sequence of acceleration \jp{in x and y} commands, $\tau_s=\{u_{s,k}\mid 1 \leq k \leq T\}$, 
We assume known dynamics $\revision{\theta}_{s, k+1} = f_s(\revision{\theta}_{s,k}, u_{s,k})$, where $\revision{\theta}_{s, k}$ is the state of robot $s$ at time $k$, and a known initial state $\revision{\theta}_{s, 0}$.
% The simulated robot trajectories consist of 2D accelerations and the real robot trajectories consist of 2D velocities.
At each timestep, we execute the first action in the trajectory and rerun the optimization, as in model predictive control (MPC).
This approach is akin to a multi-robot version of Stein MPC~\cite{lambert2020steinmpc}.

For this experiment, we assume the graph is fully-connected. We employ a loopy version of belief propagation, in which the messages are initialized and iteratively updated. This approach does not provide exact marginals but has proven to be effective in practice~\cite{murphy1999loopy}.

\textbf{Potential functions:}
The unary potential for each robot trajectory is defined with respect to the running cost $c(\revision{\theta}_{s,k}, u_{s,k})$ and terminal cost $C(\revision{\theta}_{s,T})$ for a trajectory:
\begin{equation}\label{eq:unary_factor}
    \phi_{s}(\tau_s, \revision{\theta}_{s, 0}) = \exp -\left(C(\revision{\theta}_{s,T}) + \sum_{k=1}^{T-1} \gamma_{k} \, c_s(\revision{\theta}_{s,k}, u_{s,k})\right)
\end{equation}
where $T$ is the time horizon and $\gamma_k$ are constants\revision{, and the initial state replaces the ``observation,'' $z_s$, from Eq.~\eqref{eq:bp_sum_bel}}. The running cost consists of a quadratic cost and an obstacle avoidance cost based on the signed-distance function for the obstacles. Intermediate state values $\revision{\theta}_{s,k}$ needed to compute the costs are obtained by simulated rollouts using the dynamics, $f_s(\revision{\theta}_{s,k}, u_{s,k})$.

The pairwise potential between communicating robots employs the following collision avoidance factor over the trajectory:
\begin{equation}
\label{eq:inter_robot}
\begin{split}
    \log & \; \psi_{ts}(\tau_t, \tau_s) = \\
& -\sum_{k=0}^{T} 
\begin{cases} 
      \alpha_k  \left(1 - \left( \frac{d(\revision{\theta}_{s, k}, \revision{\theta}_{t, k})}{r} \right)^\beta \right) & d(\revision{\theta}_{s, k}, \revision{\theta}_{t, k}) \leq r \\
      0 & d(\revision{\theta}_{s, k}, \revision{\theta}_{t, k}) > r \\
   \end{cases}
\end{split}
\end{equation}
where $d(\revision{\theta}_{s, k}, \revision{\theta}_{t, k})$ is the distance between the robot positions at timestep $k$, $r$ is the desired collision radius, and $\alpha_k$ and $0 < \beta \leq 1$ are constants. In our experiments, we use $r=0.5$ and $\beta=0.3$. We set $\alpha_k$ to decrease linearly over the horizon.

Given differentiable dynamics, the above potential definitions allow the gradients from Eq.~\eqref{eq:sbp_grad} to be computed with respect to the trajectories $\tau_s$. We use a Gaussian kernel which employs a distance function computed as the sum of the Euclidean distance between states in two trajectories at corresponding times. Eq.~\eqref{eq:stein-op} is applied iteratively to obtain a set of trajectories comprising the belief for each robot, $\{\tau_s^{(i)}\}_{i=1:N}$.
% We assume known, linear dynamics which allows the gradients to be computed with respect to the acceleration commands. At each timestep, we execute the first action in the trajectory and rerun the optimization, as in model predictive control (MPC).
% This approach is akin to a multi-robot version of Stein MPC~\cite{lambert2020steinmpc}.
\subsection{Baselines} 
Two baselines are employed for this scenario: the well-established Optimal Reciprocal Collision Avoidance (ORCA) algorithm~\cite{van2011reciprocal}, and Gaussian Belief Propagation (GaBP), as in~\cite{patwardhan2022distributing}. 
ORCA assumes that neighboring agent's velocity are known and calculates optimal reciprocally collision-avoiding velocities that are closest to the original preferred velocity. The scenario was implemented using the RVO2 library~\cite{RVO2Lib}. 
% for robots with radii of 20 cm and 40 cm that could achieve maximum velocities of 2 m/s. 
We assume full connectivity. 
 %Every robot was able to detect all other neighbor's velocity and would use them all for calculation.

For GaBP, potentials are expressed as a linearized Gaussian factor~\cite{davison2019futuremapping} with a bias term that encodes the expected joint Gaussian to be observed. In contrast to the formulation by Patwardhan et al.~\cite{patwardhan2022distributing}, we represent the trajectory consisting of 2D acceleration commands for one robot as a single node, rather than inferring the state at individual timesteps. We use similar potential functions to our SVBP implementation for fair comparison. 
% \jp{Jana suggestion starts here}\jm{Good suggesion. Lets add it.}
The factors in GaBP are restricted to the form:
\begin{equation}
    E_s(\tau_s) = \frac{1}{2}(h_s(\tau_s)-b_s)^\top\Sigma_s^{-1}(h_s(\tau_s)-b_s)
\end{equation}
where $h_s(\tau_s)$ is an ``observation function'' over the trajectory $\tau_s$, $b_s$ is a bias term, and $\Sigma_s$ is the covariance~\cite{davison2019futuremapping}. 

In order to use our non-linear, non-Gaussian costs, we set $h_s(\tau_s)$ to be the cost for each of our factors, with $b_s=0$.
Since our costs are non-linear, $h_s(\tau_s)$ must be linearized about an estimate via a first-order Taylor series expansion.
As in SVBP, the linearization requires backpropagation through the dynamics $f_s(\revision{\theta}_{s,k}, u_{s,k})$.
Since the quadratic cost is already linear, we use $h_s(\tau_s)=\begin{bmatrix}\revision{\Theta}_s & \tau_s\end{bmatrix}$, where $\revision{\Theta}_s$ is the state vector from simulated trajectory rollouts using the dynamics model.
Our GaBP implementation is able to infer optimal trajectories without the need of a trajectory planner by making use of the dynamics function, in contrast to the formulation by Patwardhan et al.~\cite{patwardhan2022distributing}.

\begin{figure}[t]
    \centering
    \subfloat[\label{fig:multirobot_pass_rate}]{\includegraphics[width=0.75\linewidth]{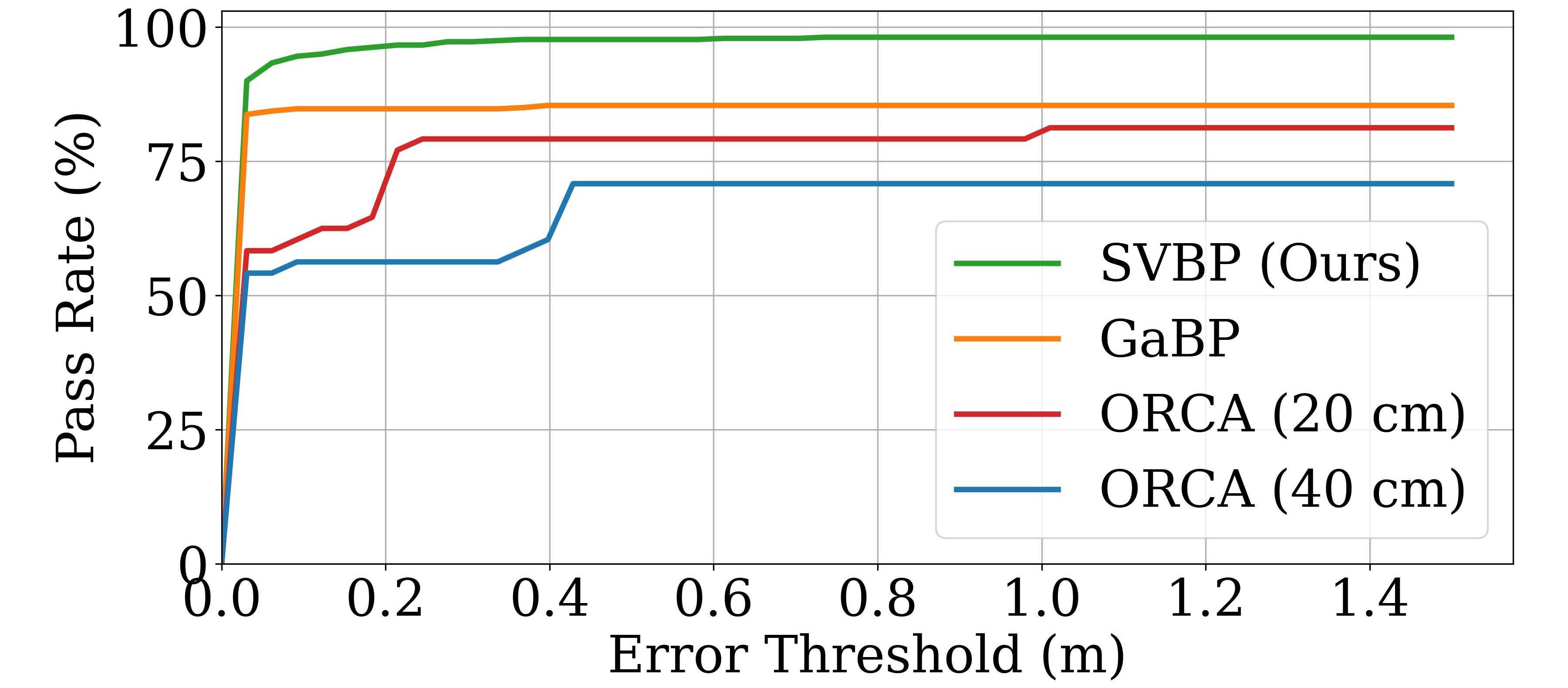}}

     % \qquad
     \subfloat[\label{fig:multirobot_planar_time}]{\includegraphics[width=0.75\linewidth]{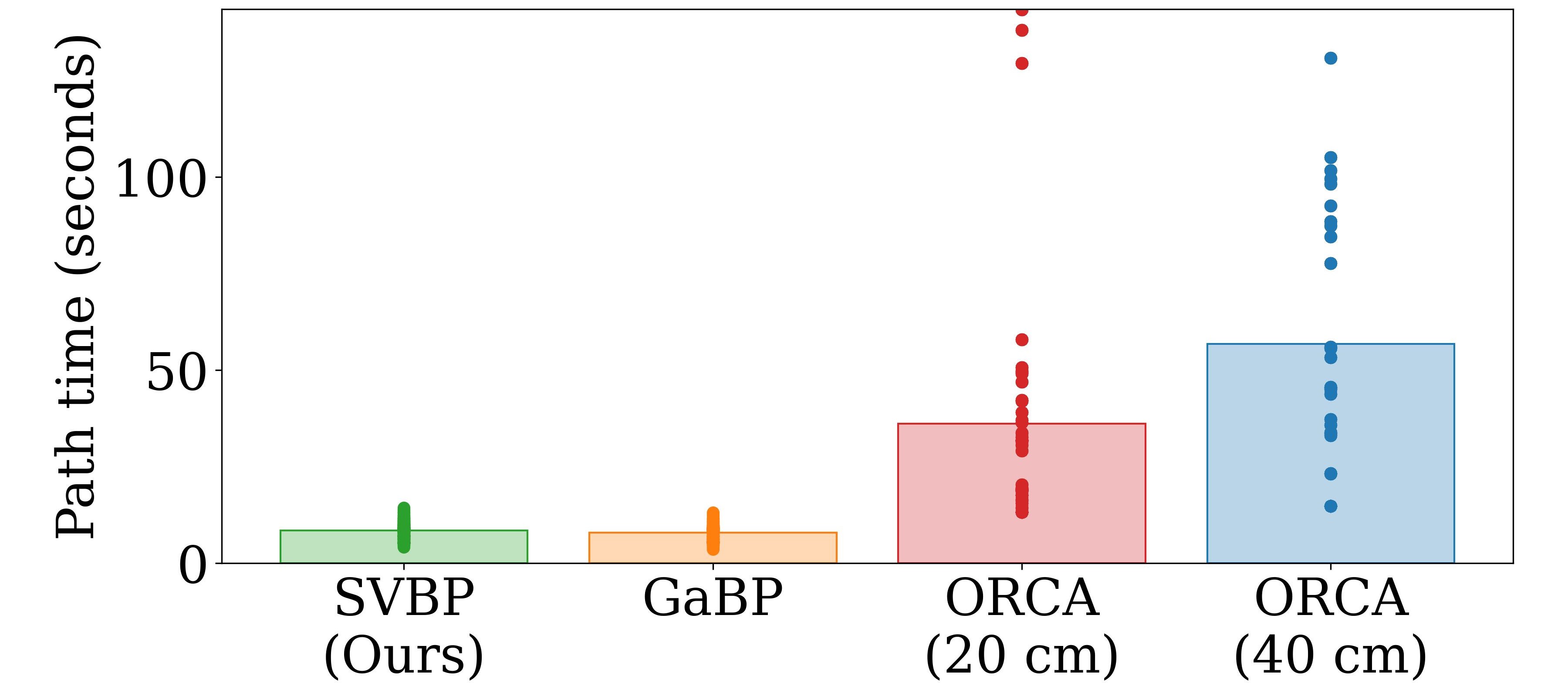}}
    \caption{Pass rate (a) and path time (b) for  each method considered for the multi-robot control example. The pass rate represents the percentage of trajectories which finished within a given error threshold. 
    Only successful results are included for path time analysis. A trajectory is successful if it reaches the goal within 30 cm without collisions.} 
\end{figure}

\subsection{Simulated Robot Experiments}

% \textbf{Implementation Details:}
We perform the simulated experiments in acceleration space, where the state $\revision{\theta}_{s, k}$ consists of 2D position and velocity, and the control commands $u_{s, k}$ are 2D accelerations. We use a time horizon of 20 discrete steps of 0.1 seconds each, making $\tau_s$ 40 dimensional for each robot. The first control command from the lowest cost trajectory\revision{, equivalent to the heighest weight particle,} is executed at each timestep. The optimization is then rerun in MPC-style.

\textbf{Results:}
We present the pass rate for ORCA, GaBP, and SVBP in \cref{fig:multirobot_pass_rate}. The pass rate represents the percentage of trajectories ($y$-axis) which reached the goal within a given error threshold ($x$-axis) across all robots for each run. Any robots that collided with another robot are not counted as passed for any threshold.
Since ORCA is sensitive to the robot radius parameter, we show results for both a radius of 20 cm and 40 cm. We perform 10 runs on each of the environments in \cref{fig:swarm_envs}. 
The total path time for each method is shown in \cref{fig:multirobot_planar_time}.
Path time is only computed for trajectories which terminated within 30 cm of the goal without collisions. While all methods result in similar path lengths, the robots move much more conservatively in the ORCA baseline, which results in higher path times. 

\begin{figure}[t]
    \centering
    \subfloat[ORCA (40 cm)]{\includegraphics[width=0.38\linewidth]{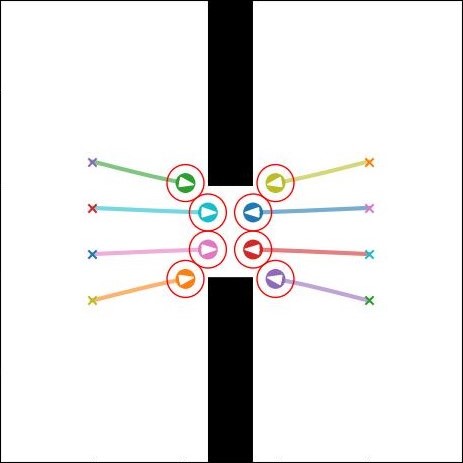}}
         \label{fig:multirobot_failures_orca}
     \qquad
     \subfloat[GaBP]{\includegraphics[width=0.38\linewidth]{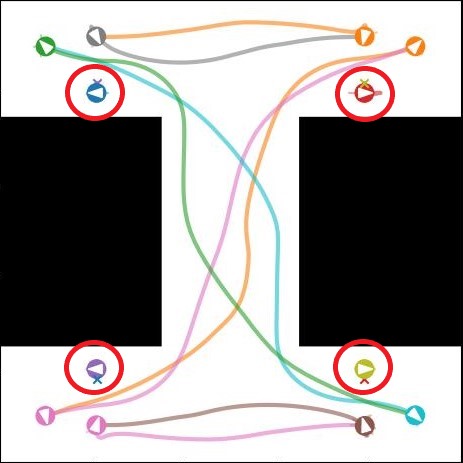}}
     \label{fig:multirobot_failures_gabp}
    \caption{Failure modes for the baselines considered for the planar navigation experiment. (a) ORCA is prone to deadlock, especially in the presence of obstacles. All run for this environment fail with a 40 cm radius (shown with red circles). (b) Gaussian Belief Propagation is prone to falling into local minima, especially around large objects. The four robots circled in red cannot get around the obstacles.} 
    \label{fig:multirobot_failures}
\end{figure}

We observe that the failure modes in SVBP can occur due to local minima, for example
% are often due to imprecision in achieving the goal location due to the tradeoff between factors. Another failure mode results when robots get stuck in local minima 
around large obstacles such as in the environments in \cref{fig:swarm_envs}(c, e). 
% Whe diverse trajectory distribution allows the robots to escape local minima in most cases. 
GaBP is especially susceptible to getting caught in local minima in the presence of challenging obstacles. A subset of robots fail to reach their goals for every run in one environment, as illustrated in \cref{fig:multirobot_failures}(b). \revision{ORCA} is particularly prone to deadlock scenarios when it must obey a collision tolerance (i.e. 40 cm collision radius case), failing for all runs in the environment shown in \cref{fig:multirobot_failures}(a).
% See \cref{fig:multirobot_failures} for an illustration of failure modes of the baselines.

\section{Real Robot Experiments}

We run our controller on a real multi-robot system comprised of omni-directional MBots~\cite{gaskell2024mbot}. We perform a collision avoidance experiment with three robots where the robots must cross paths to reach their goal locations.
The goal of this experiment is to determine the performance of our controller under 
1) noisy perception and dynamics,
2) limited computing resources,
3) realistic asynchronous message passing schedules.
The robots are equipped with a single-board computer with limited processing power (a Raspberry Pi) and pass messages through a custom websocket interface over a WiFi connection.
Each robot performs SLAM using a 2D Lidar for state estimation.
% We employ particle-based Monte-Carlo Localization using a 2D Lidar for state estimation~\cite{probrob:thrun}.
The swarm is assumed to be fully-connected. %For the real robot experiments, the control signals are 2D velocities and the states are 2D positions.

\textbf{Decentralized Message Passing:}
Each robot independently maintains a representation of the graph and updates messages within their local graph based on neighbor belief.
At the beginning of each optimization iteration, the robots request the current trajectory distribution from their neighbors which is used to update the local messages in each robot's representation of the graph.
The robots pass messages using a custom API which passes messages through websockets, inspired by {\em rosbridge}~\cite{crick2017rosbridge}, which allows them to synchronously query data from robots on a shared network.
% Each robot independently maintains and updates messages within their local graph representation, and keeps the particles representing the neighbor belief constant. 

\textbf{Baseline:} 
ORCA is implemented on the robots as a baseline. The algorithm is run in a centralized manner on a single robot which broadcasts velocity commands to the whole fleet. ORCA outputs a velocity command for each robot rather than a trajectory, therefore we do not use an external trajectory tracker and execute the velocity command directly. 
% The baseline uses the noisy state estimates from the SLAM system. 

\textbf{Implementation Details:}
% The state estimates used to roll out trajectories are subject to realistic noisy localization from the SLAM system. 
We perform the simulated experiments in velocity space, where the state $\revision{\theta}_{s, k}$ consists of 2D position and the control commands $u_{s, k}$ are 2D velocities. 
We plan over 10 discrete timesteps, with a 0.25 second timestep. We first perform 15 initialization iterations over random trajectory particles before beginning execution. 
The lowest cost trajectory is chosen and executed by a closed-loop velocity controller. 
% \jm{(Potential question: why lowest cost trajectory and not highest weighted one)} \jp{Those are the same!}
The optimization is repeated until the goal is reached in MPC-style, initializing using particles from the previous timestep.

% \subsection{Results}

\begin{figure}[t]
    \centering
    \includegraphics[width=0.75\linewidth]{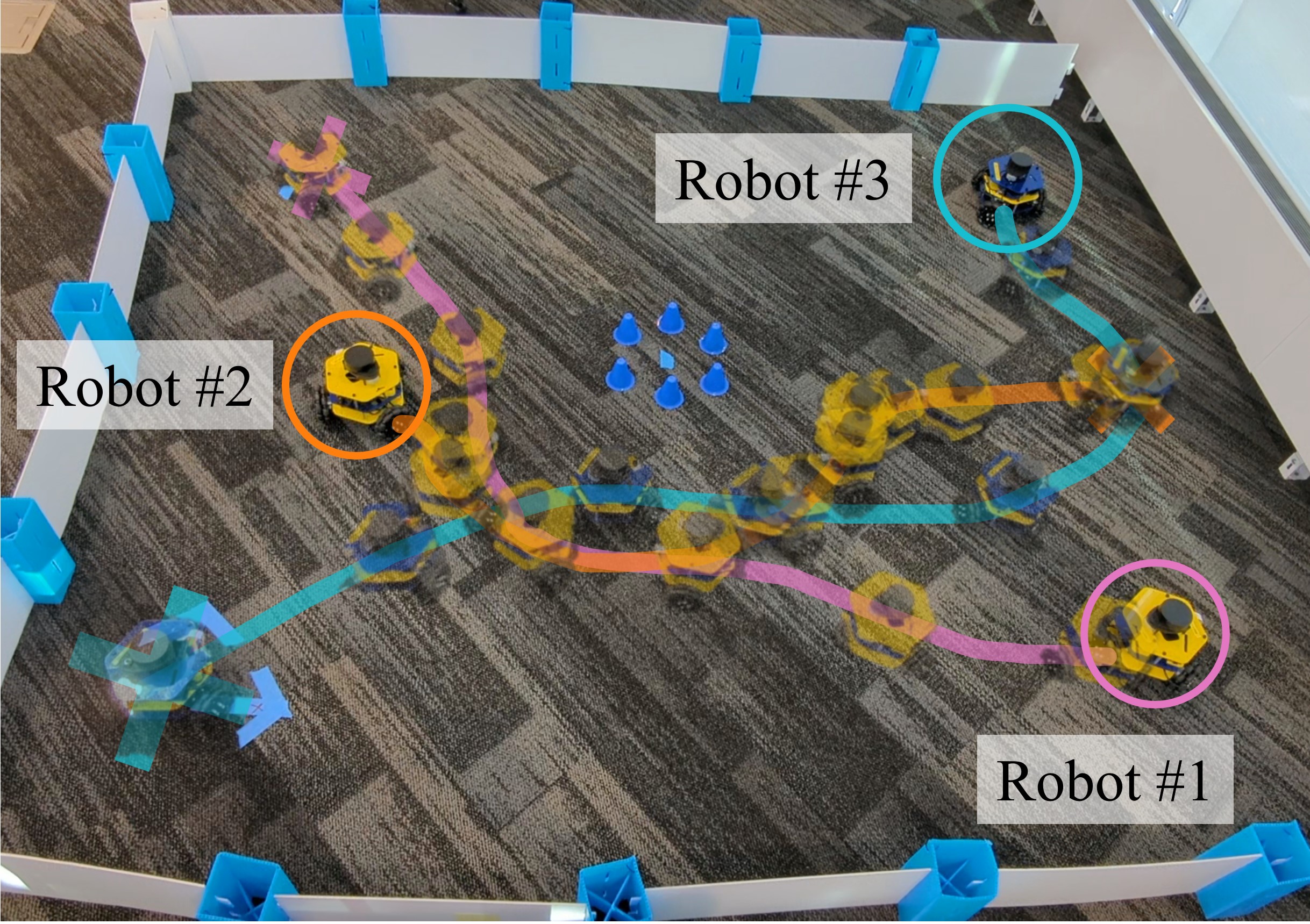}
    \caption{An example of a trajectory for the SVBP algorithm on a real robot. The circles highlight the final goal position for each robot.
    }
    \label{fig:qualitative_robot}
\end{figure}

\begin{figure}[t]
    \centering
    \includegraphics[width=0.95\linewidth]{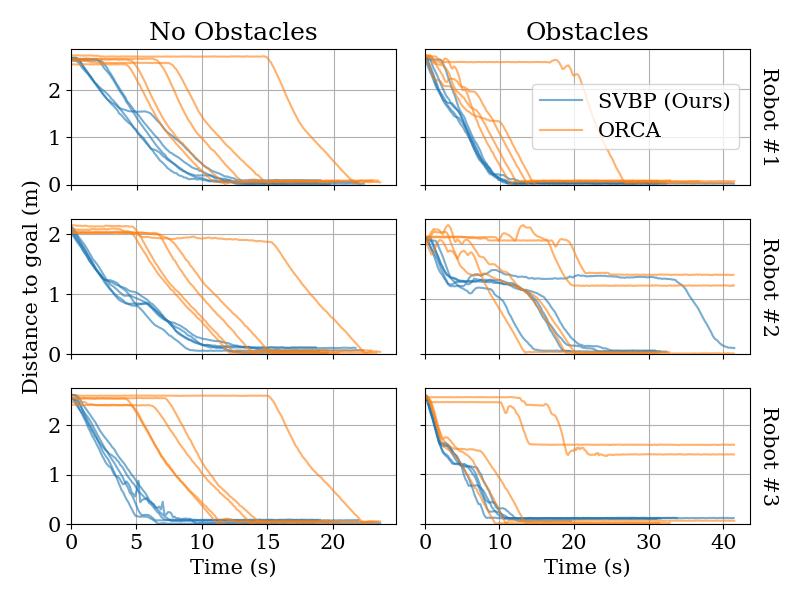}
    \caption{Distance to the goal over time for each robot for runs with no obstacles (left) and with obstacles (right).}
    \label{fig:quantitative_robot}
\end{figure}

\textbf{Results:}
We perform 5 runs on a scene with and without \revision{obstacles} (10 runs total) for SVBP and ORCA. 
% Control is then performed in MPC-style, where the best trajectory is tracked by a velocity controller while the optimization is repeated for the following timestep, initialized from the previous particles.
The time-to-goal results are shown in \cref{fig:quantitative_robot} \revision{for each of the robots shown in \cref{fig:qualitative_robot}}. 
On the scene with no obstacles, SVBP reaches the goal in all runs with no collisions except for in one run, in which one robot has a localization failure resulting in a collision.
ORCA deadlocks at the start of the trajectory for all runs. To obtain meaningful comparisons, we manually perturb the robots from their start positions to escape deadlock. ORCA's built-in random perturb for deadlock prevention fails
% is not sufficient to overcome deadlock 
in practice as
% as the physical robots require higher commanded velocities to overcome static forces, but 
ORCA tends to select low speeds which are insufficient to displace the physical robots unless large clearances are available.
Modification of the perturbation functionality for this application could mitigate this issue.
After the deadlock is resolved, ORCA and SVBP achieve similar time-to-goal in scenarios without obstacles. 

For the case with obstacles, ORCA deadlocks at the beginning of the run in two cases. The algorithm gets stuck in deadlock for 2 of 5 runs near the obstacle, and the deadlock results in a collision in one of the cases (robots \#2 and \#3 do not reach the goal in 2 cases in \cref{fig:quantitative_robot}, right). We only apply manual perturbation for deadlocks for ORCA at the beginning of the run. SVBP reaches the goals in all the cases with smoother paths. 
A visualization of an execution of SVBP with obstacles is shown in \cref{fig:qualitative_robot}. 

\section{Discussion \& Conclusion}
% DISCUSSION & CONCLUSION
We present Stein Variational Belief Propagation (SVBP), an algorithm for inferring nonparametric marginal beliefs in graphs using Stein Variational Gradient Descent. 
We demonstrate the applicability of our algorithm on two applications: simulated multi-robot perception, and multi-robot planning both in simulation and on real robots.
Through simulated perception experiments, we show that SVBP approximates the true marginal distributions better and is more particle efficient than sampling-based baselines. 
The planning experiments show that the algorithm is more effective at escaping local-minima and deadlock scenarios than baselines. The real-world planning experiments show that the method is robust to realistic noise.

% \textbf{Robot experiments:}
A limitation of the proposed algorithm is that the computation time scales with the number of neighbors.
We limited the robot experiments to three robots in order to achieve fast enough execution to run MPC on the single-board computers on the robots.
Improving the efficiency of the implementation would allow the size of the swarm to be increased.
% Future work will involve improving the implementation efficiency in order to increase the size of the swarm.
Another limitation is the need to time-synchronize incoming messages from other robots. We observe that the robots are prone to starting and stopping behavior when near other robots which we posit occurs due to lack of time synchronization. This could be mitigated by accounting for the time delays between messages. Further study is needed on the impact of message delays and packet loss.

Our robot experiments show that SVBP is robust to realistic perception and action noise, despite the fact that we do not explicitly model this noise. Integrating explicit perception and action noise models into the framework in order to deal with more challenging scenarios is an interesting avenue of investigation.

\bibliographystyle{IEEEtran}
\bibliography{ref.bib}

% Generated by IEEEtran.bst, version: 1.14 (2015/08/26)
\begin{thebibliography}{10}
\providecommand{\url}[1]{#1}
\csname url@samestyle\endcsname
\providecommand{\newblock}{\relax}
\providecommand{\bibinfo}[2]{#2}
\providecommand{\BIBentrySTDinterwordspacing}{\spaceskip=0pt\relax}
\providecommand{\BIBentryALTinterwordstretchfactor}{4}
\providecommand{\BIBentryALTinterwordspacing}{\spaceskip=\fontdimen2\font plus
\BIBentryALTinterwordstretchfactor\fontdimen3\font minus \fontdimen4\font\relax}
\providecommand{\BIBforeignlanguage}[2]{{%
\expandafter\ifx\csname l@#1\endcsname\relax
\typeout{** WARNING: IEEEtran.bst: No hyphenation pattern has been}%
\typeout{** loaded for the language `#1'. Using the pattern for}%
\typeout{** the default language instead.}%
\else
\language=\csname l@#1\endcsname
\fi
#2}}
\providecommand{\BIBdecl}{\relax}
\BIBdecl

\bibitem{fiorini1998motion}
P.~Fiorini and Z.~Shiller, ``Motion planning in dynamic environments using velocity obstacles,'' \emph{The international journal of robotics research}, vol.~17, no.~7, pp. 760--772, 1998.

\bibitem{van2011reciprocal}
J.~Van Den~Berg, S.~J. Guy, M.~Lin, and D.~Manocha, ``Reciprocal n-body collision avoidance,'' in \emph{Robotics Research: The 14th International Symposium (ISRR)}.\hskip 1em plus 0.5em minus 0.4em\relax Springer, 2011, pp. 3--19.

\bibitem{pmlr-vR4-attias03a}
H.~Attias, ``Planning by probabilistic inference,'' in \emph{International Workshop on Artificial Intelligence and Statistics}, ser. Proceedings of Machine Learning Research, vol.~R4.\hskip 1em plus 0.5em minus 0.4em\relax PMLR, 2003, pp. 9--16.

\bibitem{toussaint2009robot}
M.~Toussaint, ``Robot trajectory optimization using approximate inference,'' in \emph{International Conference on Machine Learning (ICML)}, 2009, pp. 1049--1056.

\bibitem{schwertfeger2007multi}
J.~N. Schwertfeger and O.~C. Jenkins, ``Multi-robot belief propagation for distributed robot allocation,'' in \emph{2007 IEEE 6th international conference on development and learning}.\hskip 1em plus 0.5em minus 0.4em\relax IEEE, 2007, pp. 193--198.

\bibitem{tolstaya2020learning}
E.~Tolstaya, F.~Gama, J.~Paulos, G.~Pappas, V.~Kumar, and A.~Ribeiro, ``Learning decentralized controllers for robot swarms with graph neural networks,'' in \emph{Conference on robot learning}.\hskip 1em plus 0.5em minus 0.4em\relax PMLR, 2020, pp. 671--682.

\bibitem{patwardhan2022distributing}
A.~Patwardhan, R.~Murai, and A.~J. Davison, ``Distributing collaborative multi-robot planning with {G}aussian belief propagation,'' \emph{Robotics and Automation Letters}, vol.~8, no.~2, pp. 552--559, 2022.

\bibitem{liu2016stein}
Q.~Liu and D.~Wang, ``Stein variational gradient descent: A general purpose {B}ayesian inference algorithm,'' \emph{Advances in Neural Information Processing Systems}, vol.~29, 2016.

\bibitem{sudderth2010nonparametric}
E.~B. Sudderth, A.~T. Ihler, M.~Isard, W.~T. Freeman, and A.~S. Willsky, ``Nonparametric belief propagation,'' \emph{Communications of the ACM}, vol.~53, no.~10, pp. 95--103, 2010.

\bibitem{isard2003pampas}
M.~Isard, ``{PAMPAS}: Real-valued graphical models for computer vision,'' in \emph{Computer Vision and Pattern Recognition (CVPR)}, vol.~1, 2003.

\bibitem{ihler2009particle}
A.~Ihler and D.~McAllester, ``Particle belief propagation,'' in \emph{Artificial Intelligence and Statistics}, 2009, pp. 256--263.

\bibitem{butterfield2009modeling}
J.~Butterfield, O.~C. Jenkins, D.~M. Sobel, and J.~Schwertfeger, ``Modeling aspects of theory of mind with markov random fields,'' \emph{International Journal of Social Robotics}, vol.~1, pp. 41--51, 2009.

\bibitem{gerkey2004formal}
B.~P. Gerkey and M.~J. Matari{\'c}, ``A formal analysis and taxonomy of task allocation in multi-robot systems,'' \emph{The International journal of robotics research}, vol.~23, no.~9, pp. 939--954, 2004.

\bibitem{rossi2021multi}
F.~Rossi, S.~Bandyopadhyay, M.~T. Wolf, and M.~Pavone, ``Multi-agent algorithms for collective behavior: A structural and application-focused atlas,'' \emph{arXiv preprint arXiv:2103.11067}, 2021.

\bibitem{tolstaya2021learning}
E.~Tolstaya, L.~Butler, D.~Mox, J.~Paulos, V.~Kumar, and A.~Ribeiro, ``Learning connectivity for data distribution in robot teams,'' in \emph{International Conference on Intelligent Robots and Systems (IROS)}.\hskip 1em plus 0.5em minus 0.4em\relax IEEE, 2021, pp. 413--420.

\bibitem{morgan2014model}
D.~Morgan, S.-J. Chung, and F.~Y. Hadaegh, ``Model predictive control of swarms of spacecraft using sequential convex programming,'' \emph{Journal of Guidance, Control, and Dynamics}, vol.~37, pp. 1725--1740, 2014.

\bibitem{ong2015coop}
H.~Y. Ong and J.~C. Gerdes, ``Cooperative collision avoidance via proximal message passing,'' in \emph{2015 American Control Conference (ACC)}, 2015, pp. 4124--4130.

\bibitem{van2017dmpc}
R.~Van~Parys and G.~Pipeleers, ``Distributed model predictive formation control with inter-vehicle collision avoidance,'' in \emph{2017 11th Asian Control Conference (ASCC)}, 2017, pp. 2399--2404.

\bibitem{dai2017distributed}
L.~Dai, Q.~Cao, Y.~Xia, and Y.~Gao, ``Distributed {MPC} for formation of multi-agent systems with collision avoidance and obstacle avoidance,'' \emph{Journal of the Franklin Institute}, vol. 354, no.~4, pp. 2068--2085, 2017.

\bibitem{luis2020online}
C.~E. Luis, M.~Vukosavljev, and A.~P. Schoellig, ``Online trajectory generation with distributed model predictive control for multi-robot motion planning,'' \emph{IEEE Robotics and Automation Letters}, vol.~5, no.~2, pp. 604--611, 2020.

\bibitem{murphy1999loopy}
K.~P. Murphy, Y.~Weiss, and M.~I. Jordan, ``Loopy belief propagation for approximate inference: An empirical study,'' in \emph{Conference on Uncertainty in Artificial Intelligence (UAI)}, ser. UAI'99.\hskip 1em plus 0.5em minus 0.4em\relax Morgan Kaufmann Publishers Inc., 1999, p. 467–475.

\bibitem{wainwright2008graphical}
M.~J. Wainwright, M.~I. Jordan \emph{et~al.}, ``Graphical models, exponential families, and variational inference,'' \emph{Foundations and Trends in Machine Learning}, vol.~1, no. 1--2, pp. 1--305, 2008.

\bibitem{weiss1999correctness}
Y.~Weiss and W.~Freeman, ``Correctness of belief propagation in gaussian graphical models of arbitrary topology,'' \emph{Advances in neural information processing systems}, vol.~12, 1999.

\bibitem{davison2019futuremapping}
A.~J. Davison and J.~Ortiz, ``{FutureMapping} 2: {G}aussian belief propagation for spatial {AI},'' \emph{arXiv preprint arXiv:1910.14139}, 2019.

\bibitem{murai2022robot}
R.~Murai, J.~Ortiz, S.~Saeedi, P.~H. Kelly, and A.~J. Davison, ``A robot web for distributed many-device localisation,'' \emph{arXiv preprint arXiv:2202.03314}, 2022.

\bibitem{Desingh2019pmpnbp}
K.~Desingh, S.~Lu, A.~Opipari, and O.~C. Jenkins, ``Efficient nonparametric belief propagation for pose estimation and manipulation of articulated objects,'' \emph{Science Robotics}, vol.~4, no.~30, 2019.

\bibitem{pavlasek2020parts}
J.~Pavlasek, S.~Lewis, K.~Desingh, and O.~C. Jenkins, ``Parts-based articulated object localization in clutter using belief propagation,'' in \emph{International Conference on Intelligent Robots and Systems ({IROS})}.\hskip 1em plus 0.5em minus 0.4em\relax IEEE, 2020.

\bibitem{Opipari:2023:DNBP:3592762}
A.~Opipari, J.~Pavlasek, C.~Chen, S.~Wang, K.~Desingh, and O.~C. Jenkins, ``{DNBP}: Differentiable nonparametric belief propagation,'' \emph{ACM/IMS Journal of Data Science}, vol.~1, no.~1, 2024.

\bibitem{lienart2015epbp}
T.~Lienart, Y.~W. Teh, and A.~Doucet, ``Expectation particle belief propagation,'' in \emph{Advances in Neural Information Processing Systems}, 2015, pp. 3609--3617.

\bibitem{pacheco2014dpmp}
J.~Pacheco, S.~Zuffi, M.~Black, and E.~Sudderth, ``Preserving modes and messages via diverse particle selection,'' in \emph{International Conference on Machine Learning (ICML)}, vol.~32, no.~2, 2014, pp. 1152--1160.

\bibitem{liu2016ksd}
Q.~Liu, J.~D. Lee, and M.~Jordan, ``A kernelized {S}tein discrepancy for goodness-of-fit tests and model evaluation,'' in \emph{International Conference on Machine Learning (ICML)}, 2016.

\bibitem{lambert2020steinmpc}
A.~Lambert, A.~Fishman, D.~Fox, B.~Boots, and F.~Ramos, ``Stein variational model predictive control,'' in \emph{Conference on Robot Learning (CoRL)}, 2020.

\bibitem{barcelos2021dual}
L.~Barcelos, A.~Lambert, R.~Oliveira, P.~Borges, B.~Boots, and F.~Ramos, ``Dual online {S}tein variational inference for control and dynamics,'' in \emph{Robotics: Science and Systems ({RSS})}, 2021.

\bibitem{9661415}
F.~A. Maken, F.~Ramos, and L.~Ott, ``{S}tein {ICP} for uncertainty estimation in point cloud matching,'' \emph{Robotics and Automation Letters}, vol.~7, no.~2, pp. 1063--1070, 2022.

\bibitem{9811656}
A.~Lambert, B.~Hou, R.~Scalise, S.~S. Srinivasa, and B.~Boots, ``Stein variational probabilistic roadmaps,'' in \emph{International Conference on Robotics and Automation (ICRA)}.\hskip 1em plus 0.5em minus 0.4em\relax IEEE, 2022, pp. 11\,094--11\,101.

\bibitem{wang2018stein}
D.~Wang, Z.~Zeng, and Q.~Liu, ``Stein variational message passing for continuous graphical models,'' in \emph{International Conference on Machine Learning}.\hskip 1em plus 0.5em minus 0.4em\relax PMLR, 2018, pp. 5219--5227.

\bibitem{zhuo2018message}
J.~Zhuo, C.~Liu, J.~Shi, J.~Zhu, N.~Chen, and B.~Zhang, ``Message passing {S}tein variational gradient descent,'' in \emph{International Conference on Machine Learning}.\hskip 1em plus 0.5em minus 0.4em\relax PMLR, 2018, pp. 6018--6027.

\bibitem{casella1992explaining}
G.~Casella and E.~I. George, ``Explaining the {G}ibbs sampler,'' \emph{The American Statistician}, vol.~46, no.~3, pp. 167--174, 1992.

\bibitem{gretton2012kernel}
A.~Gretton, K.~M. Borgwardt, M.~J. Rasch, B.~Sch{\"o}lkopf, and A.~Smola, ``A kernel two-sample test,'' \emph{The Journal of Machine Learning Research}, vol.~13, no.~1, pp. 723--773, 2012.

\bibitem{garreau2017large}
D.~Garreau, W.~Jitkrittum, and M.~Kanagawa, ``Large sample analysis of the median heuristic,'' \emph{arXiv preprint arXiv:1707.07269}, 2017.

\bibitem{RVO2Lib}
\BIBentryALTinterwordspacing
J.~Van Den~Berg, S.~J. Guy, J.~Snape, M.~Lin, and D.~Manocha. (2016) {RVO2} library: Reciprocal collision avoidance for real-time multi-agent simulation. [Online]. Available: \url{https://gamma.cs.unc.edu/ORCA/}
\BIBentrySTDinterwordspacing

\bibitem{gaskell2024mbot}
P.~Gaskell, J.~Pavlasek, T.~Gao, A.~Narula, S.~Lewis, and O.~C. Jenkins, ``{MBot}: A modular ecosystem for scalable robotics education,'' in \emph{International Conference on Robotics and Automation ({ICRA})}, 2024.

\bibitem{crick2017rosbridge}
C.~Crick, G.~Jay, S.~Osentoski, B.~Pitzer, and O.~C. Jenkins, ``{\em {r}osbridge}: {ROS} for non-{ROS} users,'' in \emph{Robotics Research: The 15th International Symposium ISRR}.\hskip 1em plus 0.5em minus 0.4em\relax Springer, 2017, pp. 493--504.

\end{thebibliography}

\end{document}